\documentclass{article}
\usepackage[nonatbib,final]{neurips_2022}

\usepackage[utf8]{inputenc} 
\usepackage[T1]{fontenc}    
\usepackage{hyperref}       
\usepackage{url}            
\usepackage{MnSymbol}       
\usepackage{booktabs}       
\usepackage{amsfonts}       
\usepackage{nicefrac}       
\usepackage{microtype}      
\usepackage{xcolor}         
\usepackage{graphicx}
\usepackage{amsmath}
\usepackage{biblatex}
\bibliography{source}

\begin{document}

\title{\textit{DeepRepViz}: Identifying Confounders in Deep Learning Predictions}

\author{
  Roshan Prakash Rane \thanks{\href{https://psychiatrie-psychotherapie.charite.de/en/metas/contactform/adresse/roshan_prakash_rane/x}{\url{https://psychiatrie-psychotherapie.charite.de/en/metas/contactform/adresse/roshan_prakash_rane/}}} \\
  Department of Psychology, Humboldt-Universität zu Berlin, \\ 
  Charité – Universitätsmedizin Berlin, Einstein Center for Neurosciences Berlin \\
   \And
   JiHoon Kim \\
   Department of Education and Psychology\\
   Freie Universität Berlin \\
   \And
   Arjun Umesha \\
   Independent \\
   \And
   Didem Stark \\
   Department of Psychiatry and Neurosciences\\
   Charité – Universitätsmedizin Berlin \\
   \And
   Marc-André Schulz \\
   Department of Psychiatry and Neurosciences\\
   Charité – Universitätsmedizin Berlin \\
   \And
   Kerstin Ritter \\
   Department of Psychiatry and Neurosciences\\
   Charité – Universitätsmedizin Berlin \\
}

\maketitle




\begin{abstract} Deep Learning (DL) models have gained popularity in neuroimaging studies for predicting psychological behaviors, cognitive traits, and brain pathologies. However, these models can be biased by confounders such as age, sex, or imaging artifacts from the acquisition process. 
To address this, we introduce ‘DeepRepViz’, a two-part framework designed to identify confounders in DL model predictions. The first component is a visualization tool that can be used to qualitatively examine the final latent representation of the DL model. The second component is a metric called ‘Con-score’ that quantifies the confounder risk associated with a variable, using the final latent representation of the DL model. We demonstrate the effectiveness of the Con-score using a simple simulated setup by iteratively altering the strength of a simulated confounder and observing the corresponding change in the Con-score.
Next, we validate the DeepRepViz framework on a large-scale neuroimaging dataset (n=12000) by performing three MRI-phenotype prediction tasks that include (a) predicting chronic alcohol users, (b) classifying participant sex, and (c) predicting performance speed on a cognitive task called ‘trail making’. DeepRepViz identifies sex as a significant confounder in the DL model predicting chronic alcohol users (Con-score=0.35) and age as a confounder in the model predicting cognitive task performance (Con-score=0.3). In conclusion, the DeepRepViz framework provides a systematic approach to test for potential confounders such as age, sex, and imaging artifacts and improves the transparency of DL models for neuroimaging studies. \end{abstract}

\section*{Introduction}

Deep Learning (DL) offers a promising avenue for neuroimaging research \cite{vieira17_deepni-review, ritter21_psyreview} as it can be trained directly on high-dimensional ($p>>n$) neuroimaging modalities, such as structural Magnetic Resonance Imaging (MRI), functional Magnetic Resonance Imaging (fMRI), and Electroencephalogram (EEG).
Particularly in population neuroscience research \cite{paus10_popneuro}, DL models have been employed to understand the relationship between the brain and various psychological phenotypes \cite{sui20_MLinPsy}, as well as brain pathologies \cite{sui20_MLinPsy, ritter21_psyreview, vieira17_deepni-review}.
For example, DL models are trained on brain MRI to predict the risk of developing Alzheimer’s disease \cite{khojaste22_DLandADNI} or to identify subtypes in psychiatric disorders such as major depressive disorder or substance use disorders \cite{drysdale17_rsMRIdepression, rane22_aim}.
However, instead of learning the brain-phenotype relationship relevant to the research, DL models can learn to use demographic factors or spurious biases encoded in the neuroimaging data  \cite{chen23_biasinMedAI}. Such extraneous factors are known as ‘confounders’.
For instance in Thibeau et al. (2022) \cite{thibeau22_adniscanner}, their DL model used to predict the development of Alzheimer's disease was found to rely on a spurious bias induced by the MRI acquisition process \cite{thibeau22_adniscanner}. In their dataset, the majority of the participants in the dataset that developed Alzheimer’s were scanned with a 3 Tesla MRI scanner, while the majority of controls were scanned with a 1.5 Tesla scanner \cite{thibeau22_adniscanner}. The DL model predicted all participants scanned with a 3 Tesla scanner as having a risk of developing Alzheimer’s disease, rather than using biomarkers in the brain to predict Alzheimer’s \cite{chen23_biasinMedAI}. Such a `confounded' model is not clinically useful.
Similarly, since higher brain atrophies associated with Alzheimer’s are also more prevalent among older people \cite{hua10_ADNIsexage}, age can also be a confounder when predicting Alzheimer’s disease. If a model simply classifies all older participants as the Alzheimer’s disease group without taking into consideration the underlying brain biomarkers, then such a model would not be clinically valid either \cite{polsterl23_causalAD}.
Several studies routinely report such confounders affecting their model predictions \cite{hyatt20_covarreview, abbasi20_reviewskin, seyyed21_chestxraybias, thibeau22_adniscanner}  \cite{chen23_biasinMedAI,ritter21_psyreview}.
As confounders are highly prevalent in observational neuroimaging datasets, there is a pressing need in the field for a framework to detect confounded DL models.

To avoid confounded models, neuroimaging studies typically choose 2 to 5 variables as potential confounders and control for them \cite{hyatt20_covarreview}. These generally include age, sex, intracranial brain volume, educational level, and socio-economic status of the participants \cite{hyatt20_covarreview}. 
The variables are controlled using confound control methods such as regressing out the confounder \cite{snoek19_control, johnson07_combat}, or counterbalancing the data \cite{rane22_imagen, rane23_eating}. Controlling for confounders can prevent confounders from influencing model predictions. 
Interestingly, the confounders that are controlled in a study can often exert a greater influence on model performance than the choice of the modeling algorithm \cite{rane22_imagen} or the input neuroimaging modality \cite{komeyer24_confound2D}. 
In addition to the commonly chosen potential confounders, other confounders may arise from the specific research question \cite{pearl16_causal} or biases induced by the dataset’s sampling and image acquisition process \cite{gorgen18_SAA}. 
Therefore, selecting a fixed set of confounders without considering the specific research question or dataset biases may lead to a suboptimal study design \cite{polsterl23_causalAD, hyatt20_covarreview}.
To prevent unexpected confounders from leaking into the modeling stage, Görgen et al. 2018 \cite{gorgen18_SAA} recommend systematically assessing all variables in a study for their `confounder risk' using the Same Analysis Approach.
Building on this recommendation, we propose the `\textit{DeepRepViz}' framework for predictive DL models. The \textit{DeepRepViz} framework offers a setup for systematically assessing all variables in a study for their `confounder risk' by inspecting the final latent representation learned by DL models \cite{bengio13_representation}.  Our proposed framework comprises 2 components:
\begin{enumerate}   
    \item A web-based visualization tool (\url{https://deep-rep-viz.vercel.app}) to visually inspect the representation learned by the DL model in its final layer. The tool aids in qualitatively identifying confounded predictions.
    \item A metric, `Con-score', that quantifies the confounder risk associated with various potential confounders of the study.
\end{enumerate}

In the following sections, we will outline the theoretical basis of \textit{DeepRepViz} and the Con-score metric. We will then demonstrate the efficacy of the \textit{DeepRepViz} framework on three neuroimaging-based brain-phenotype prediction tasks.

\section*{Method} 

In a predictive model $f_{model}: X \mapsto y$, a confounder variable $c^k$ is defined as an extraneous variable encoded in the exposure $X$ and associated with the outcome $y$ \cite{snoek19_control}. For example, in Thibeau et al.'s work (2022) \cite{thibeau22_adniscanner}, the MRI scanner strength was a confounder as it affected their MRI and correlated with the Alzheimer's disease diagnosis due to a systematic data acquisition bias. A confounder variable $c^k$ introduces an alternative predictive pathway $X \leftmapsto c^k \mapsto y$ \cite{pearl16_causal}. If the predictive model $f_{model}$ uses this confound-driven pathway for predicting $y$, then $f_{model}$ is said to be `confounded' by $c^k$ \cite{pearl16_causal}. In Thibeau et al.'s case, their model exploited the spurious confounder pathway (MRI$ \leftmapsto $scanner$ \mapsto$Alzheimer) rather than relying on biomarkers of the Alzheimer's disease in the MRI (MRI$\mapsto$Alzheimer). 

To estimate the confounder risk associated with $c^k$ on the prediction task $f_{model}: X \mapsto y$, we can measure the strength of the confounder-driven pathway $X \leftmapsto c^k \mapsto y$ \cite{pearl16_causal, rane22_imagen}. In this study, we propose such a confounder risk estimation method called `Con-score' that can be used with predictive DL models. 

\begin{figure}[htbp]
\thispagestyle{empty}
    \centering
    \includegraphics[width=0.9\textwidth]{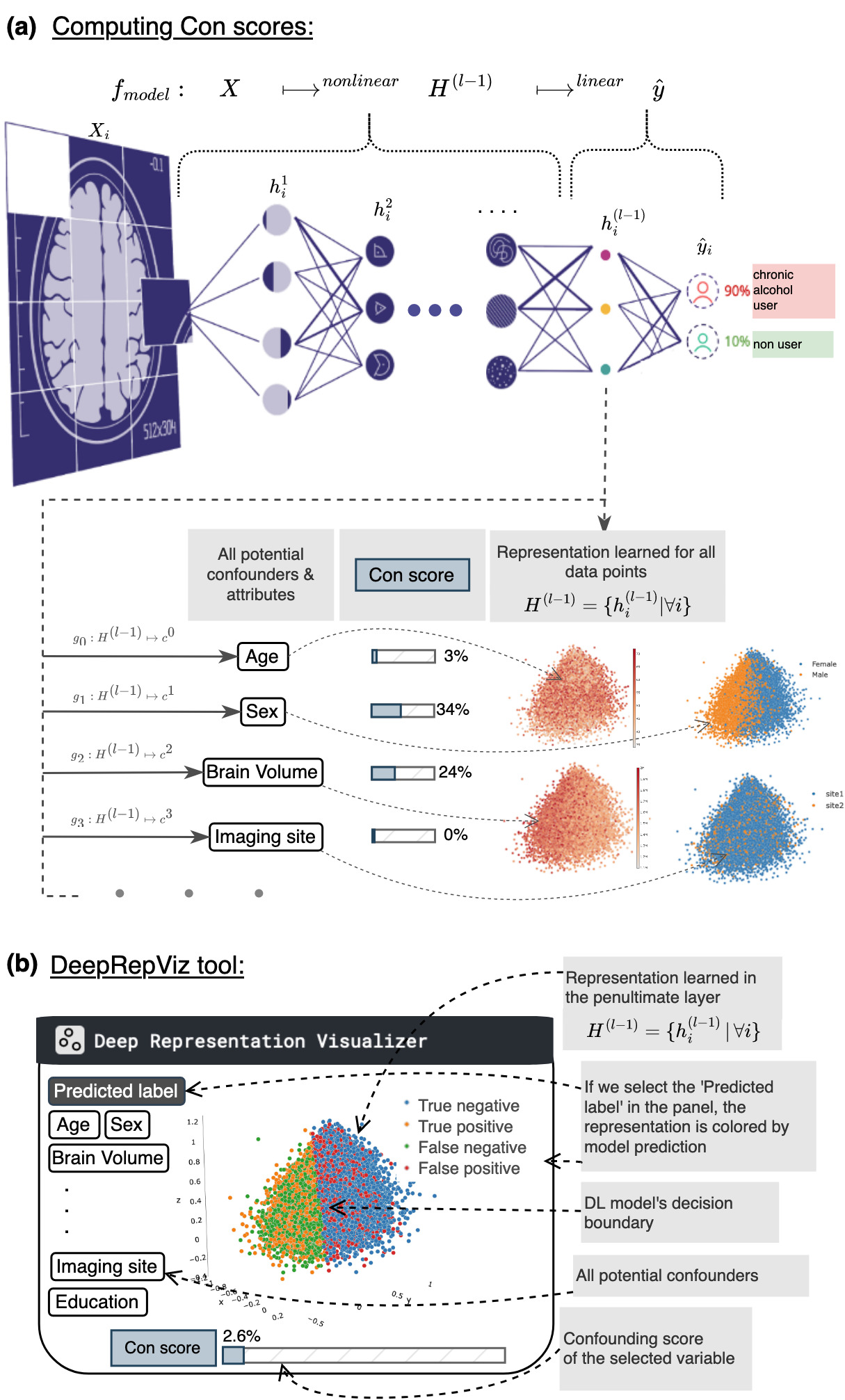}
    \thispagestyle{empty}
    \caption{\textbf{A demonstration of \textit{DeepRepViz} and the Con-score metric:} Figure (a) shows a DL model that classifies chronic alcohol users from non-users using the participants' structural MRI data. Con-score is computed for variables in the data using the representation learned in the penultimate layer $H^{(l-1)}$ of the DL model. Con-score is highest for `sex' when classifying chronic alcohol users. Figure(b) shows the \textit{DeepRepViz} tool and how it can be used to inspect the learned representation $H^{(l-1)}$. When we select the predicted label $\hat{y}$ in the tool, we can see the linear decision boundary of the model in $H^{(l-1)}$. This decision boundary aligns with the representation of sex in $H^{(l-1)}$ in Figure(a). This is also reflected in the Con-score. This implies that the model could be using the information about the participant's sex encoded in the MRI data as a proxy to identify chronic alcohol users.}\label{fig:overview}
\end{figure}

\textbf{Con-score derivation:} From the perspective of representation learning theory\cite{bengio13_representation}, a DL model can be imagined to comprise two steps as demonstrated in \autoref{fig:overview}(a): $X \mapsto^{nonlinear} H^{(l-1)} \mapsto^{linear} y$. The model first transforms the data into a condensed latent representation using a series of non-linear layers ($H^{l-1}$), and then the final layer generates the predictions by linearly mapping from $H^{(l-1)}$ to $y$. We propose estimating the confounder risk using the final representation $H^{(l-1)}$. If we assume that the DL model uses the alternative confounder pathway $X \leftmapsto c^k \mapsto y$ for making its predictions, then we would expect two things to be true: first, $c^k$ would be linearly predictable from $H^{l-1}$ (i.e., $X \mapsto H^{l-1} \mapsto c^k$ exists), and second, the linear model predicting $c^k$ ($f_{c^k}:H^{l-1} \mapsto c^k$) would correspond with the final layer of the DL model predicting $y$ ($f_{model}^{(l-1)}:H^{l-1} \mapsto y$). We develop the Con-score metric by combining these two criteria. Therefore, the Con-score metric which is given in \autoref{eq:con_score}:

\begin{equation} \label{eq:con_score}
    \text{Con-score}\,(c^k) = \: R_{c^k}^2 \cdot \,|\, \cos\,(\theta_{c y}) 
\end{equation}

Here, $R_{c^k}^2$ is the coefficient of multiple determination \cite{nagelkerke1991_R2} of the linear model predicting $c^k$ ($f_{c^k}:H^{l-1} \mapsto c^k$), and $\cos(\theta_{c y})$ is the cosine similarity between the linear model predicting $c^k$ and the final DL layer predicting $y$. If $c^k$ is categorical, then we use McKelvey and Zavoina's $pseudo\,R^2$ \cite{mckelvey1975_pseudoR2} instead of $R^2$. The $\theta_{c y}$ in the second term is obtained by taking the vector angle between the parameter of the linear model $f_{c^k}:H^{l-1} \mapsto c^k$ and the parameters of the DL layer performing $f_{model}^{(l-1)}:H^{l-1} \mapsto y$. The final Con-score ranges between $[0,1]$. The higher the Con-score, the higher the likelihood that the model is using the alternative (confounder) pathway $H^{l-1} \leftmapsto c^k \mapsto y$ for its prediction. A Con-score of 1 indicates that the DL model has learned all information about $c_k$ in $H^{l-1}$, and that the linear prediction of $y$ and the linear prediction of $c_k$ are exactly the same. 

\textbf{\textit{DeepRepViz} visualization tool:} We can simultaneously estimate the Con-score for all variables associated with the neuroimaging prediction task, such as the MRI scanning parameters, demographic, and socio-economic factors \cite{gorgen18_SAA, rane23_eating}. We offer an interactive web-based tool to qualitatively inspect the latent representation of the DL model $H^{l-1}$ and contrast it with the Con-score for all the variables. If the Con-score is high for $c^k$, then we can expect $c^k$ to be clustered in the representation space $H^{l-1}$ \cite{glocker21_splitconf} and aligned with the prediction of $y$. \autoref{fig:overview} demonstrates how the tool can be used to identify confounders using an example of a DL model predicting chronic alcohol use. Currently, the tool requires the representation space to be 3-dimensional $H^{(l-1)} \in \mathbb{R}^3$. However, the Con-score metric generalizes to representations of any arbitrary dimension $H^{(l-1)}\in \mathbb{R}^n$. 
Apart from the Con-score, the tool also computes other metrics for comparison, such as the Silhouette Coefficient for categorical variables and distance correlation scores for continuous variables. Please refer to the tool documentation for more information\footnote{\url{https://deep-rep-viz.vercel.app/docs.html}}. 

\textbf{Experiment design:} 
We begin by testing the Con-score metric on simulated data to assess its performance under controlled settings with different boundary conditions. We then evaluate the utility of the metric and the visualization tool on three brain-phenotype prediction tasks using neuroimaging data.

The simulated dataset contains a binary label $y$, a binary confounder $c$, and a 2-dimensional input data $H=\{h_0,h_1\}$. We generate eight instances of the dataset by systematically altering the correlation between $y$, $c$, and the input data $H$. In the figures, different colors represent the binary states of the label $y$, while different shapes denote the binary states of the confounder $c$. The top row instances (numbered 1 to 4) are generated such that $c$ can be easily predicted with a linear classification model from the input features $H$. Whereas in the bottom row instances (5 to 8), $c$ classification becomes relatively difficult. As we move from instance number 1 to 4 or from 5 to 8, we incrementally change the correlation between $c$ and $y$. For instance, in instance number 1, $c$ and $y$ are completely uncorrelated, while in instance 3, $c$ and $y$ become completely correlated.

On a subsample of $n=12000$ from the UK Biobank dataset, we conduct three exemplary brain-phenotype prediction tasks using a state-of-the-art DL architecture, 3D ResNet-50. We predict the participant's (1) alcohol use, (2) sex, and (3) performance at a cognitive task using the T1-weighted structural MRI data. In the first task, we classify chronic alcohol users from non-users of alcohol. In the second task, we predict the sex of participants from their structural MRI data. In the third task, we predict the time taken by the participants to complete the `trail-making’ cognitive test.

\section*{Results}
\begin{figure}[h!]
    \includegraphics[width=1.\textwidth]{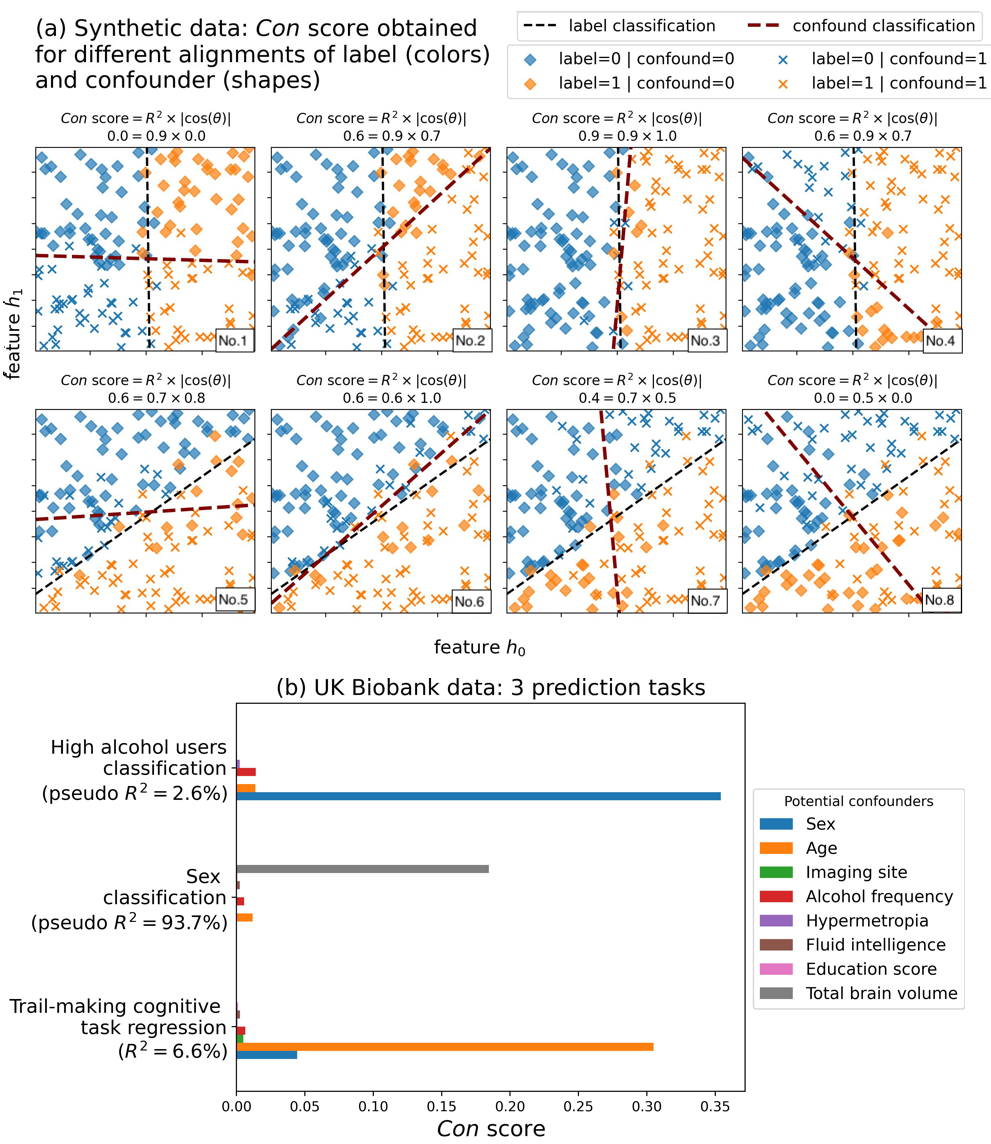}
    \caption{\textbf{Results of applying Con-score metric to (a) simulated dataset and (b) UK Biobank neuroimaging dataset}:
    (a) shows Con-scores obtained on a simulated binary classification task for 8 different levels of correlations between a confounder $c$, a binary label $y$, and the input features $H=\{h_0, h_1\}$.
    (b) shows the Con-scores obtained for eight potential confounder variables (see legend) in three brain-phenotype prediction tasks on the UK Biobank dataset.} \label{fig:results}
\end{figure}

The results of the experiments on both a simulated dataset and a neuroimaging dataset are presented in \autoref{fig:results}. The results on the simulated dataset show that as we systematically change the similarity between a confounder and the label, the Con-score changes proportionately, as expected. Experiments on the neuroimaging dataset reveal different confounders affecting the three DL model prediction tasks.

\textbf{Results on the simulated data:} \autoref{fig:results} (a) shows the Con-score obtained for the eight instances of the simulated dataset generated with different settings of the confounder $c$, a binary label $y$, and the input features $H=\{h_0, h_1\}$. The Con-scores are highest in the instance when the label classification boundary (black line) and the confound classification boundary (red line) align. In the top row, this occurs in instance 3 and in the bottom row, this occurs in instance 6. In these instances, $c$ is highly correlated with the label $y$. The second term in the Con-score metric, $\cos(\theta)$, captures this correlation as seen in each instance's title in \autoref{fig:results}(a). 
It is easier to linearly predict $c$ in dataset instances on the top row (numbers 1 to 4) compared to the bottom row. The $R^2$ term of the Con-score captures this variation,  as shown in the title of each instance in \autoref{fig:results}(a). This is clearly evident when we compare the $R^2$ of instance 3 with 6 or instance 1 with 8. In summary, the Con-scores are highest in the dataset instances where the correlation between $c$ and $y$ is high and $c$ is encoded in the exposure $H$. For all other cases, the Con-score drops down proportionately. 

\textbf{Results on neuroimaging data:} \autoref{fig:results} (b) shows the results obtained for the three brain-phenotype prediction tasks. For all the tasks, the Con-score is computed for eight variables from the UK Biobank datasets as listed in the figure legend. When classifying high alcohol users, the DL model only achieves pseudo-$R^2 = 2.6\%$ and the Con-score is highest for the sex variable ($0.35$). When we visualize the representation $H^{(l-1)}$ on \textit{DeepRepViz} (refer to \autoref{fig:overview}), we find that the majority of the participants predicted as high alcohol users by the DL model are men. This reveals that the model is picking up on the sex bias present in the data. 
For the cognitive performance prediction task, the DL model also achieves a low $R^2 = 6.6\%$  and the Con-score is highest for age ($0.3$). This suggests that the model tends to predict older participants as taking longer to complete the trail-making cognitive test. Visualization on the \textit{DeepRepViz} tool also confirms that age is encoded in the latent representation and is aligned with the model's prediction of the label. Interestingly, hypermetropia or long-sightedness was not a confounder for this task, although one can expect that good eyesight is important for such visual cognitive tests. 
For sex prediction, the Con-scores reveal that information related to the `total brain volume' of the participants is encoded in the final learned representation layer (Con-score $=0.18$) but none of the other 7 variables are encoded. Whether total brain volume should be considered as a confounder or an explanation depends on the research question behind predicting sex from the brain MRI data.

\section*{Discussion and Conclusion}

Confounders can pose a significant challenge when using predictive modeling techniques such as DL in population neuroscience research. As the size of observational neuroimaging datasets continues to grow, the issue of confounders is only going to worsen \cite{hyatt20_covarreview}. To address this challenge, we present the `\textit{DeepRepViz}' framework.

The `\textit{DeepRepViz}' framework is designed to detect confounders in DL model predictions by examining the latent representation learned by the model in its last layer. It comprises two components: a web-based visualization tool to inspect the final latent representation of the DL model, and a metric called `Con-score' that quantifies the confounder risk of a potential variable on the DL model predictions.

Using the \textit{DeepRepViz} framework in combination with predictive DL models offers several benefits. Firstly, it enables researchers to compare multiple variables and assess their impact on the representation learned by their model \cite{gorgen18_SAA}. This helps researchers to comprehend their model decisions \cite{rane23_eating} in terms of several human-understandable `concepts' \cite{kim18_TCAV}. For instance in one of our experiments, \textit{DeepRepViz} revealed that total brain volume is a crucial feature for the DL model predicting sex from brain MRI. Such a tool is especially useful for population neuroscience studies because psychological phenotypes often co-occur with various demographic, socioeconomic, and environmental factors \cite{rane23_eating}.
Secondly, once potential confounders are identified, researchers can employ the \textit{DeepRepViz} framework to test various control control methods. A successful control control method should significantly reduce the Con-score, bringing it closer to 0 \cite{rane22_imagen}.
Lastly, the tool not only allows researchers to qualitatively examine the latent representation for signs of confounders, but it also enables them to search for errors resulting from incorrect model configuration and optimization. Please refer to the tool documentation for more information \footnote{\label{doc}\url{https://deep-rep-viz.vercel.app/docs.html}}.

Finally, in population neuroscience studies using predictive models, confounder identification and confound control should not be viewed as a single step in the analysis. Rather, they should be considered an iterative process that helps us uncover clinically relevant brain-phenotype relationships with low effect sizes using machine learning \cite{rane22_imagen}. We present \textit{DeepRepViz} as a versatile tool for iteratively validating DL models, opening pathways for medical discovery from neuroimaging data with DL. Researchers can apply \textit{DeepRepViz} not only to neuroimaging but also to any other exploratory research using predictive DL models. The tool is publicly available at \url{https://deep-rep-viz.vercel.app}, and instructions on how to use it are provided in the tool documentation$^{\ref{doc}}$.

\printbibliography
\section*{Acknowledgements and Funding}
This work was funded by the DeSBi Research Unit (DFG; KI-FOR 5363; Project ID 459422098), the consortium SFB/TRR 265 Losing and Regaining Control over Drug Intake (DFG; Project ID 402170461), FONDA (DFG; SFB 1404; Project ID: 414984028) and FOR 5187 (DFG; Project ID: 442075332).
\end{document}